# Towards Active Robotic Vision in Agriculture: A Deep Learning Approach to Visual Servoing in Occluded and Unstructured Protected Cropping Environments

Paul Zapotezny-Anderson* Chris Lehnert**

*Queensland University of Technology, Brisbane, Australia
(e-mail: p.zapotezny-anderson@connect.qut.edu.au).
** Queensland University of Technology, Brisbane, Australia
(e-mail:c.lehnert@qut.edu.au)

Abstract: 3D Move To See (3DMTS) is a mutli-perspective visual servoing method for unstructured and occluded environments, like that encountered in robotic crop harvesting. This paper presents a deep learning method, Deep-3DMTS for creating a single-perspective approach for 3DMTS through the use of a Convolutional Neural Network (CNN). The novel method is developed and validated via simulation against the standard 3DMTS approach. The Deep-3DMTS approach is shown to have performance equivalent to the standard 3DMTS baseline in guiding the end effector of a robotic arm to improve the view of occluded fruit (sweet peppers): end effector final position within 11.4 mm of the baseline; and an increase in fruit size in the image by a factor of 17.8 compared to the baseline of 16.8 (avg.).

## 1. INTRODUCTION

Robotic harvesting is gaining greater importance as the harvesting of high-value crops such as apples, citrus and sweet peppers (capsicums) is labour intensive and becoming less economically viable due to the increasing price of skilled labour and decreasing availability (Gongal *et al.*, 2015).

The challenge of dealing with occlusions (e.g. concealing leaves or branches) in agricultural robotics is commonly reported and is considered as a non-trivial technical issue for which further research is warranted (Gongal *et al.*, 2015). In contrast to the environment of a typical industrial robot, i.e. a factory where the environment is well structured, a farming environment is unstructured, cluttered and heavily occluded. In a greenhouse, each plant and their fruits are unique in their shape and structure, and the location of fruit to harvest is not known *a priori*.

A common approach to resolve the issue of occlusions is to use visual servoing via template matching of the target object with the occluded object in the image and use the matched template to determine the pose of the occluded target object (Chen *et al.*, 2011). However, many crops come in a broad variety of shapes which makes the use of templates problematic.

An example of a robotic harvesting platform is "Harvey" (Fig. 1), an agricultural robot designed to harvest sweet peppers autonomously (Perez *et al.*, 2015; McCool *et al.*, 2016; Sa *et al.*, 2017; Lehnert *et al.*, 2017). *Harvey* is a state-of-the-art concept that has been successfully tested in real protected cropping environments (Lehnert *et al.*, 2017). Despite *Harvey's* success in harvesting sweet peppers under test conditions, *Harvey* is challenged when the sweet peppers are occluded by leaves.

In 2018, Lehnert *et al.* proposed the 3D Move To See (3DMTS) approach to guide the end effector of the robotic arm around occlusions to obtain an uncluttered view of the target fruit in preparation for robotic harvesting. The approach did not rely on templates of the fruit or other elements of the cropping environment *a priori*. The 3DMTS approach (Fig. 1) guides the end effector in the direction to optimise the amount of revealed fruit in the image, while accounting for the robot arm's mobility, by comparing images taken from nine imaging sensors arranged in a 3D array on the end effector.

The 3DMTS approach was initially tested and refined under a simulation environment and was then tested on realistic replicas of sweet pepper plants using *Harvey* as the test platform. The 3DMTS approach was shown to be able to guide the end effector around occlusions to obtain a better view of the fruit and presented a three-fold increase in performance (improved view of the fruit) in comparison to a baseline method. However, the 3DMTS implementation suffers a low visual servoing rate of approximately 1 Hz due to the large amount of imaging data to process and limitations of the supporting hardware architecture; this impacts the practicability of 3DMTS approach for potential utilisation in robotic fruit harvesting. Lehnert *et al.* (2018) identified future work in investigating methods to reduce the number of imaging sensors by possibly estimating the direction of the gradient from a single imaging sensor using a deep learning technique that uses data captured from the 3D imaging sensor array as training data. A reduction in the number of sensors would also reduce the complexity of the supporting hardware architecture. Overall, changes to the amount of sensor data, visual servoing approach and hardware requirements could increase the visual servoing rate.

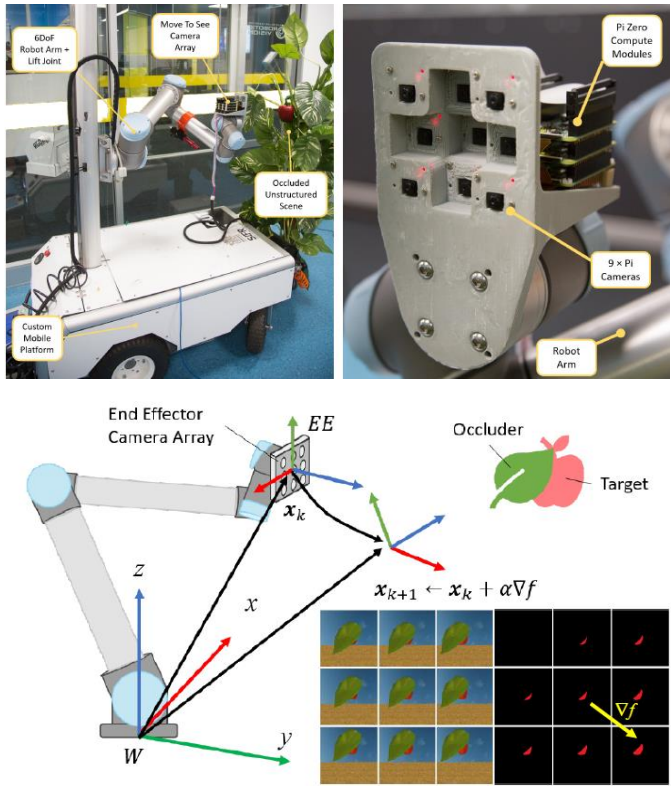

Fig. 1. *Harvey* configured for 3DMTS (top left). 3D camera array (top right). 3DMTS concept (bottom): The camera array captures multiple images of the occluded target fruit to determine a direction gradient, $\nabla f(x)$, based on the proportion of pixels attributed to the target fruit in the images. The direction gradient is used to update the current position, $x_k$, of the end-effector, where $x_k$ is the translation vector between the world reference frame, *W*, and the end effector reference frame, *EE* (Lehnert *et al.*, 2018).

The research presented here follows on from prior work of Lehnert *et al.* (2018) to investigate approaches for 3DMTS that could reduce the number of imaging sensors needed (i.e. less than nine, but preferably one) to guide the end effector of a robotic arm around occlusions to better view a target object in a time efficient manner. The contributions of this paper are a deep learning approach for single-camera 3DMTS that is based on a CNN, and the validation of the approach in a realistic simulation environment. This research is significant because it aims to advance the practical feasibility of 3DMTS, an approach that readily addresses occlusions in unstructured environments, such as agricultural settings, for which image templates cannot be applied.

## 2. LITERATURE REVIEW

This section provides a review of the literature of methods that have the potential to improve the time efficiency of 3DMTS, while reducing the number of imaging sensors required. These approaches include reformulating the optimisation strategy, direct exploitation of image geometry and deep learning. Each approach is assessed for suitability, with assessment outcomes summarised in the concluding subsection.

### 2.1 Reformulating the Optimisation Strategy

The 3DMTS approach (Lehnert *et al.*, 2018), shown in Fig. 1, uses local (one-step ahead) optimisation of an objective function to determine the direction to improve the view of the target fruit, where an improved view is taken to be a view in which the fruit occupies a greater proportion of the image; i.e. a less occluded view. A one-step ahead strategy has been taken because the objective function cannot be modelled over the configuration space of the harvester *a priori* due to the unstructured nature of the environment (plants and fruit). The values of the objective function must be determined *in situ* from measurements. The objective function, $f(x)$, is defined as

$$f(x) = w_1 p(x) + w_2 m(x), \text{ given } w_1 + w_2 = 1 \quad (1)$$

where $x$ is the state of the end effector, $p(x)$ is the proportion of target object pixels in the image obtained via an image segmentation process, and $m(x)$ is the measure of the robot arm's mobility, with $w_1$ and $w_2$ as the respective weights.

The objective function is determined locally in a "one-shot" approach via images taken by nine cameras in a 3D array of known geometry. The benefit of the "one-shot" approach is that the local objective function is determined in a single step; it does not need to be developed over a series of steps, and so is suitable for dynamic environments. However, it achieves this at the cost of the time to process the large amount of imagery data concurrently, which limits the visual servoing rate to 1 Hz – a rate that is not practicable.

Reducing the number of cameras reduces the burden of data processing and management but it also reduces the fidelity of the local objective function under a one-shot approach. It is assumed that using a single camera will adversely impact the accuracy in estimating the direction to take to get the next best view of the fruit. Using an optimisation method with a single camera requires the objective function to be sampled via a series of measurements over several time steps. Such a strategy is less suitable to dynamic environments as the value of the objective function at a location becomes time dependent.

Under a naïve strategy (Roy *et al.*, 2004), the sensor can be moved in a predefined scanning path to obtain multiple views of the target object to determine the local value of the objective function before determining the optimal move towards achieving the final objective. Furthermore, this process is repeated for every optimal control decision. The total number of sensor moves is high and the trajectory of the sensor is convoluted. The strategy is expensive in time and energy. Probing strategies (Farrokhsiar *et al.*, 2013) can be employed to apply scanning explorations more effectively, reducing the number scanning attempts throughout the engagement. The trajectory of the sensor is sub-optimal but converges over the course of the engagement.

Frew (2003) employs an optimal dual control strategy which requires only a single measurement to be made at each optimal control step. Here, platform acceleration control decisions for manoeuvring are made to minimise the amount of searching (scanning moves) required to achieve the final objective in the smallest number of steps. Sensor trajectories are quite direct, with a small lateral deviation from the path to the final objective. Such a strategy for single-sensor 3DMTS would involve a control decision (manoeuvre) to maximise the local objective function while minimising the uncertainty of its directional gradient. The strategy could be implemented via an Extended Kalman Filter if the dynamics can be appropriately linearised, but would require computationally expensive numerical methods. Reducing the number of imaging sensors becomes a dual problem between the time to process image data and the time needed to scan or explore the environment. While there are strategies that reduce the time to scan, they increase implementation complexity and risk, and would struggle in a dynamic environment.

*2.2 Direct Exploitation of Image Geometry*

Exploiting the scene geometry to determine the direction to move would obviate the need to determine and optimise the 3DMTS objective function, and hence the need for the 3D sensor array. 3DMTS utilises image segmentation to identify regions of the image attributable to the target fruit, and with modification may be able to identify leaves. However, further information would be required to determine if identified leaves, or parts thereof, are occluding or not occluding. Depth information of the imaged scene would help identify occlusions. Depth information may be obtained by viewing the scene from multiple viewpoints, or via augmenting the imaging sensor, e.g. a colour-depth (RGB-D) sensor (Gongal *et al.*, 2015).

Using a single imaging sensor to view the scene from multiple viewpoints faces the same challenges and pitfalls as described above for obtaining objective function values. Next Best View strategies from the field of 3D-scanning can minimize the scanning path (either distance travelled, or number of views used) to obtain a complete model of a target object with complex, self-occluding geometries (Chen *et al.*, 2011). However, such strategies are geared to achieve total coverage of the target object or scene, not merely sufficient coverage as might be needed in the 3DMTS case. Alternatively, multiple viewpoints can be achieved via stereo-vision techniques. Font *et al.* (2014) utilised stereovision to identify and locate fruit to grasp. Depth of the fruit was determined by triangulating the centroids obtained via image segmentation of the image from each perspective, though such a technique was not able to determine the location of occlusions. This would require depth measurement per pixel via stereo-matching which is computationally expensive, and when done with a short baseline between imaging sensors, such as the case with 3DMTS, often yields depth estimates with high uncertainty (Gongal *et al.*, 2015).

Tanigaki *et al.* (2008) and Sa *et al.* (2017) both used a depth augmented imaging sensor to obtain a 3D point cloud of the scene around the target fruit (cherries and sweet peppers, respectively) which was coupled with image segmentation techniques to identify regions attributed to fruit, stems and leaves. In both cases, the geometry was exploited to locate cutting sites for harvesting. As suggested in Chen *et al.* (2011), such 3D models might be used to match template models of the target fruit to estimate the target fruit pose for visual servoing. However, as discussed in depth in Lehnert *et al.* (2018), many harvestable crops are quite varied in shape and cannot be sufficiently represented by a model. Nonetheless, the 3D point cloud with image segmentation might be used to identify boundaries of the exposed fruit that represent occlusions and determine a direction to move the sensor to improve the view of the fruit as per 3DMTS; this has been identified as a future research question to address.

*2.3 Deep Learning*

Deep learning has become the preferred technique for analysing images, according to a survey conducted by Kamilaris and Prenafeta-Boldú (2018), with trends shifting from Support Vector Machine classifiers to CNNs. The attraction to CNNs is driven by automated feature extraction which obviates manual feature engineering and its superior performance in comparison to other machine learning techniques. While CNNs require longer training times and larger training sets, testing times are short. Morrision *et al.* (2018) applied a CNN for grasping objects in a cluttered environment where many occlusions were present. Overcoming the issue of occlusion in the clutter required multiple sensor viewpoints to be considered. The CNN used depth images to select the sensor viewpoints to improve visibility for visual grasp detection of miscellaneous and adversarial objects located in a cluttered tray. Models or templates of the objects were not used to identify objects or grasping locations. The system achieved an 80% success rate with approximately 10 seconds per grasp attempt. The system operated at a rate of 10 Hz; that is ten times the rate of 3DMTS. This example of a CNN matches the profile of single-sensor 3DMTS well: single imaging sensor, no models or templates, cluttered environment with occlusions, determination of a final pose with best visibility for grasping, and a high processing rate. A CNN trained for single sensor 3DMTS could use the sensor image to determine the direction to the next best view.

*2.4 Summary and Implications*

The three approaches have been assessed for suitability to improve the time efficiency of 3DMTS in guiding an end effector around occlusions to better view target fruit. Of the three approaches, the deep learning approach appears to provide the best opportunity at a lower risk as prior work suggests that it best matches the profile of the problem. The application of a CNN to 3DMTS addresses the need for effective visual servoing in unstructured and occluded environments, and so has been selected for development and evaluation by simulation.

3. DEVELOPMENT OF DEEP-3DMTS

The development goal of the CNN is to generate a direction gradient for end effector position updates from an image of a single camera. As shown in Fig. 2, the CNN will render the peripheral cameras of the 3DMTS sensor array redundant and replace the image processing and objective function optimisation stages of standard 3DMTS, which is expected to improve the visual servoing rate of 3DMTS. The CNN can be trained using the images from the reference camera (*image*$^{ref}$) and the direction gradient ($\nabla f$) using the prior 3DMTS methodology. As the mobility measure requires including joint state information into the CNN, we focused on using images only and not including mobility criteria in the objective function. Python 3.7 with PyTorch 1.1.0 was used for the development of Deep 3DMTS.

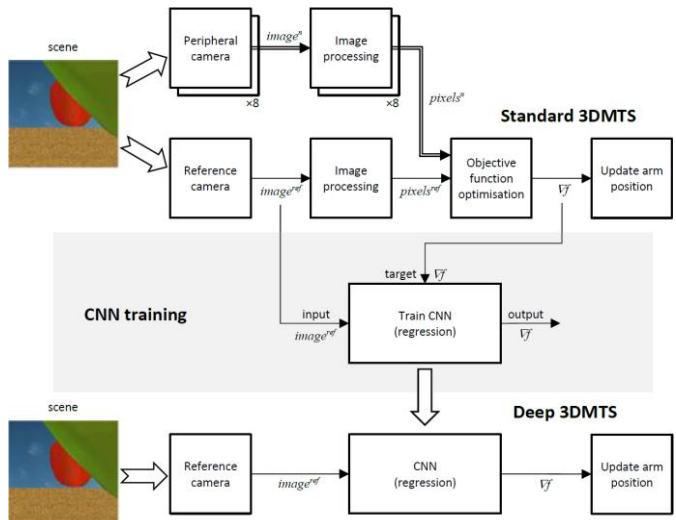

Fig. 2. Implementing a CNN for 3DMTS.

### 3.1 Generating Training and Validation Data

The data required to train a CNN includes the imagery from the central reference camera of the 3D camera array and the directional gradient at each time-step of a 3DMTS end effector trajectory. The training data was obtained from a V-REP (Rohmer *et al.*, 2012) simulation environment established by Lehnert *et al.* (2018) to develop 3DMTS. The simulations were run with the occluding leaf set at random positions and orientations, as conducted under the conditions detailed in Table 1. A total of 55 eligible trajectories where recorded including reference camera images with associated direction gradient target values; simulation runs in which the fruit was fully occluded or not occluded were excluded, as these do not warrant 3DMTS. The collected images with target values were randomly distributed (70:30) to the training and validation sets, which consists of 1155 and 495 items, respectively.

**Table 1. Simulation run parameters**

| Parameter | Value |
|---|---|
| Initial end effector position [$x, y, z$] | [0.04, 0.59, 0.68] m |
| Fruit position [$x, y, z$] | [0.4, 0.6, 0.7] m |
| Occlusion reference position [$x, y, z$] | [0.3, 0.55, 0.7] m |
| Occlusion random vert. offset range | [-0.06, 0.06] m |
| Occlusion random horiz. offset range | [-0.08, 0.08] m |
| Occlusion random angle offset range | [$-\pi/2, \pi/2$] rad |
| Objective function pixel weight, $w_1$ | 1.0 |
| Objective function mobility weight, $w_2$ | 0.0 |
| Camera array radius | 0.07 m |

### 3.2 CNN Fine-Tuning

A pre-trained ResNet18 model had its fully-connected layer modified to support the regression of three output values for the direction gradient: $\nabla f_x$, $\nabla f_y$ and $\nabla f_z$. Input images were resized and normalised for CNN training. The application of random translational image jitter for the training set was also explored to improve the robustness of the CNN. The CNN model was fine-tuned with the weights of all layers updated, as the artificial images obtained from V-REP that are used for training are a departure from the natural images used in pre-training the ResNet18 model. Fine-tuning was performed over a range of hyperparameters (Table 2), with performance assessed using the Mean Square Error (MSE) between predicted and actual target values. The CNN used for the simulation experiment had a training loss of 0.552 and validation loss of 1.896 (Fig. 3) under the fine-tuning conditions in Table 2.

**Table 2. CNN fine-tuning conditions**

| Hyperparameter | Selected Value | Search Range |
|---|---|---|
| Epochs | 50 | [25 – 100] |
| Mini-batch size | 64 | [40 – 128] |
| Learning rate | $1\times10^{-5}$, decay: 0.01 / 25 epochs | [$1\times10^{-3} - 1\times10^{-7}$] |
| Random image jitter | 0 pixels | [0 – 10 pixels] |

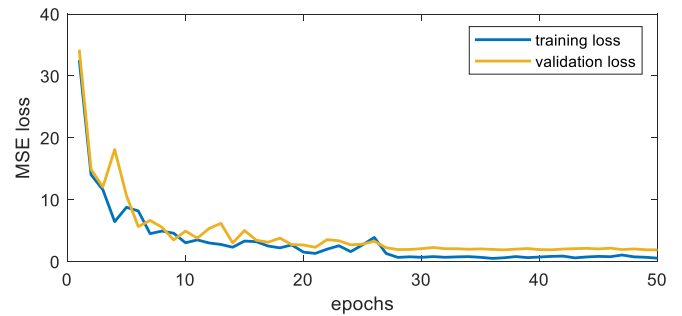

Fig. 3. MSE loss during fine-tuning of the CNN used for simulation experiments: final training loss = 0.552; final validation loss = 1.896.

### 3.3 Integrating the CNN into 3DMTS

To integrate the fine-tuned CNN into the 3DMTS process, images from the reference camera were resized and normalised prior to input to the CNN. The output of the CNN, the direction gradient, was then used to update the position of the end effector (see Fig. 2). While the image processing stages were circumvented for the determination of the direction gradient, they were still retained for two reasons: the proportion of fruit pixels in the image were used as a performance metric for comparison to standard 3DMTS; and, the difference between the centroid of the fruit pixels and the centre of the image was used to update the sensor orientation to ensure that the sensor remains pointing at the fruit throughout the engagement. Accommodating the sensor orientation updates within the CNN approach to entirely remove dependency on the 3DMTS image processing stage remains a point of investigation in future work.

## 4. EXPERIMENT AND RESULTS

The V-REP simulations used in Lehnert *et al.* (2018) were used to evaluate the effectiveness of the proposed Deep-3DMTS in comparison to standard 3DMTS (baseline method). In the experiment, the end effector of a modelled *Harvey* robot was placed in front of the target fruit (sweet pepper) which was partially occluded by a leaf set in a random pose in front of the fruit (see Table 1 for initial conditions). The end effector was then guided around the occluding leaf to provide an occlusion free view of the target fruit using the deep learning based method and the baseline method. Guidance was terminated when the magnitude of the direction gradient was sufficiently small (<1.5) or the proportion of fruit pixels in the image is large (>40%). The expectation was that guidance provided by the Deep-3DMTS approach would be similar to that provided by the baseline method. Hence, the performance metrics include the comparison of the Deep-3DMTS approach with the baseline method for: the number of guidance steps to achieve the final position; the difference in final position; and the increase of the fruit size (pixels in the image) between the start and final images as a measure of an improved view of the initially occluded fruit (see Table 3).

Two series of 12 trials were performed. Series 1 had the end effector starting at the initial position detailed in Table 1 to assess how the Deep-3DMTS approach compared with the baseline method under the same conditions used for CNN development. Series 2 included a random offset in the range of [-0.05, 0.05] m to the *x*, *y* and *z* components of initial starting position in Table 1 to assess how robust the Deep-3DMTS approach was to variation from the conditions in which it was developed, as might occur in a real-world implementation.

In all trials of each series, the Deep-3DMTS approach was able to guide the end effector around the occluding leaf to provide an occlusion free view of the fruit. A summary of performance metrics for Series 1 and 2 are presented in Table 3, which includes examples of trials from Series 1. A comparison of trajectories produced by the Deep-3DMTS approach and the baseline method for the example trials are presented in Fig. 4 as the end effector was guided to improve the view of the initially occluded fruit. The trajectories of the Deep-3DMTS approach for the trials of Series 2 are presented in Fig. 5 to illustrate the variation in starting position and the difference in end position from that of the baseline method. Fig. 5 also provides the plot of the improvement of fruit size over the course of each trajectory generated by the Deep-3DMTS approach in Series 2.

**Table 3. Performance of Deep-3DMTS with comparison against Standard 3DMTS baseline**

|  | Guidance steps | | Final position $\Delta_{BL}$ [mm] | Fruit size [% image] | | |
|---|---|---|---|---|---|---|
|  | Deep 3DMTS | $\Delta_{BL}$ |  | Start size | Deep 3DMTS | $\Delta_{BL}$ |
| Series 1 | | | | | | |
| mean | 33.4 | -3.7 | 11.1 | 1.75 | 30.59 | 1.39 |
| max | 41 | 5 | 28.7 | 3.13 | 35.46 | 5.01 |
| min | 23 | -12 | 1.8 | 0.64 | 28.76 | 0.10 |
| Ex 1 | 41 | -3 | 6.3 | 2.87 | 30.94 | 1.62 |
| Ex 2 | 33 | -1 | 1.8 | 1.57 | 30.65 | 1.18 |
| Ex 3 | 32 | -1 | 19.5 | 1.80 | 31.71 | 2.16 |
| Ex 4 | 23 | -12 | 10.5 | 0.74 | 30.53 | 1.30 |
| Series 2 | | | | | | |
| mean | 31.4 | -4.8 | 11.4 | 1.69 | 30.19 | 1.65 |
| max | 41 | 6 | 19.5 | 3.42 | 31.79 | 2.82 |
| min | 20 | -19 | 3.4 | 0.81 | 28.81 | 0.38 |

$\Delta_{BL}$ = difference of Deep-3DMTS from baseline method.

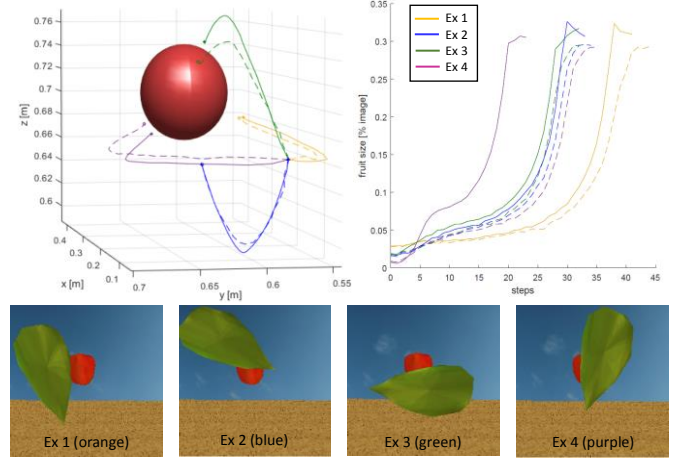

Fig. 4. Examples from Series 1: (top left) trajectories for Deep-3DMTS (solid line) and baseline method (dashed line) with respect to the fruit (red sphere) – occlusions are not shown; (top right) plot of fruit size over course of trajectories; (bottom) starting images of occluded fruit.

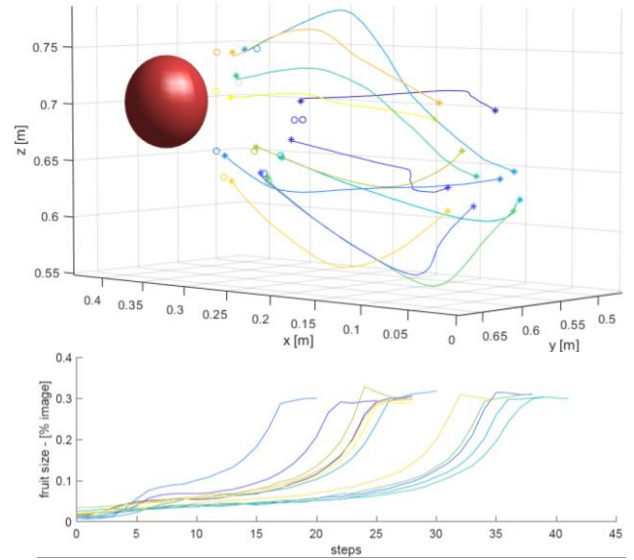

Fig. 5. Series 2 results: (top) trajectories for Deep-3DMTS with baseline method end positions (circles) with respect to the fruit (red sphere) – occlusions are not shown; (bottom) plot of fruit size over course of trajectories.

## 5. DISCUSSION AND CONCLUSION

This paper presents Deep-3DMTS, a deep learning visual servoing approach to handling occlusions for applications in agriculture. The Deep-3DMTS method uses CNN's to reduce the multi-camera 3DMTS method to a single camera approach. Overall, the Deep-3DMTS method was demonstrated in simulation to improve the view of an occluded fruit achieving similar performances to the standard 3DMTS (baseline method). On average, when random offsets were applied to the starting location, the Deep-3DMTS approach was able to arrive within 11.4 mm of the baseline method's end point. This improved the view of the fruit by increasing its size in the image, from 1.69% of the image when occluded, to 30.19% for the final occlusion-free view (increase by factor of 17.8), which is within 1.65 percentage points of the baseline. The proposed approach was sufficiently robust to a random offset to the starting position (tested to 50mm to *x*, *y* and *z* components) from which CNN training data was collected, with negligible change in performance when compared to trials where the random offset was not applied (-0.3 mm for end position and -0.40 percentage points for fruit size). The Deep-3DMTS approach often used less steps than the baseline method: 4.8 steps less on average, with 19 steps less in the extreme. This occurred when the fruit was heavily occluded. Under such conditions, the baseline method guided the end effector laterally to reduce the level of occlusion and then moved towards the fruit, whereas the Deep-3DMTS approach did both concurrently producing a shorter trajectory that used less steps (Fig. 6). Once an occlusion-free view was attained, both methods travelled the same path towards the fruit resulting in similar stopping positions and view of the fruit.

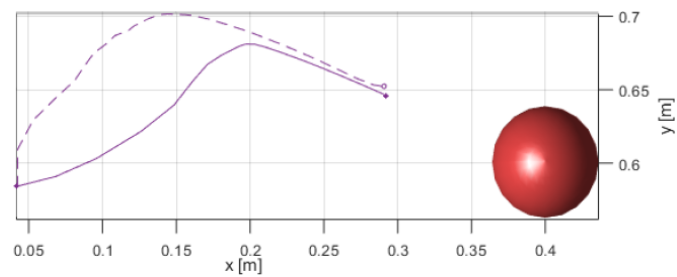

Fig. 6. Variation in trajectory resulting in less steps for Deep-3DMTS (Deep-3DMTS: solid line, baseline method: dashed-line, fruit: red sphere, occlusion: not shown).

The Deep-3DMTS approach proposed in this paper has been shown, via simulation, to have similar performance to standard 3DMTS for guiding an end effector around occlusions in an unstructured and unmodelled environment. Hence, it is feasible for a single-camera 3DMTS approach that may improve the time-efficiency of standard 3DMTS through reductions in required hardware and data processing. Future work will look into further validating the Deep-3DMTS approach in a real protected cropping environment using a robotic platform such as *Harvey* and evaluated for performance and time-efficiency.